\documentclass[runningheads]{llncs}
\usepackage[T1]{fontenc}
\usepackage{graphicx}
\usepackage{amsmath}
\usepackage{amssymb}
\usepackage{booktabs}
\usepackage{xcolor}
\usepackage{multirow}
\usepackage{hyperref}
\usepackage{float}
\usepackage{placeins}

\definecolor{cvprblue}{rgb}{0.21,0.49,0.74}
\hypersetup{
    colorlinks=true,
    hypertexnames=false,
    linkcolor=cvprblue,
    filecolor=cvprblue,
    urlcolor=cvprblue,
    citecolor=cvprblue
}

\begin{document}
\title{PRISM: Color-Stratified Point Cloud Sampling}
\titlerunning{PRISM}
\author{Hansol Lim\inst{1,2} \
Minhyeok Im\inst{3,4} \
Jongseong Brad Choi\inst{1,2,3}\thanks{Corresponding author.}}
\authorrunning{H. Lim et al.}
\institute{Department of Mechanical Engineering, State University of New York, Korea, Incheon, South Korea 
\and
Department of Mechanical Engineering, State University of New York, Stony Brook, NY, United States 
\and
Department of Computer Science, State University of New York, Korea, Incheon, South Korea 
\and
Department of Computer Science, State University of New York, Stony Brook, NY, United States}
\maketitle              
\begin{abstract}
We present PRISM, a novel color-guided stratified sampling method for RGB-LiDAR point clouds. Our approach is motivated by the observation that unique scene features often exhibit chromatic diversity while repetitive, redundant features are homogenous in color. Conventional downsampling methods (Random Sampling, Voxel Grid, Normal Space Sampling) enforce spatial uniformity while ignoring this photometric content. In contrast, PRISM allocates sampling density proportional to chromatic diversity. By treating RGB color space as the stratification domain and imposing a maximum capacity $k$ per color bin, the method preserves texture-rich regions with high color variation while substantially reducing visually homogeneous surfaces. This shifts the sampling space from spatial coverage to visual complexity to produce sparser point clouds that retain essential features for 3D reconstruction tasks.

\end{abstract}
\section{Introduction}
\label{sec:intro}

Point clouds are the foundation of modern 3D reconstruction. Traditionally, point clouds were acquired through LiDAR sensors~\cite{zhang2014loam,geiger2013kitti,roriz2022automotive}. Alternatively, image-based point cloud generation via Structure-from-Motion (SfM) offers a complementary approach~\cite{schonberger2016sfm}. Recently, sensor fusion techniques have emerged that combine LiDAR with camera data for multimodal point cloud generation~\cite{zou2025uavmm3d,palladin2025samfusion,wang2025lidarvggt}.

Multimodal sensing has become increasingly accessible. LiDAR sensor prices have dropped significantly in recent years, driven by advances in solid-state technology and mass production capabilities~\cite{liang2024evolution}. The technology has become affordable enough to be integrated into consumer smartphones and handheld scanners. The fusion of calibrated cameras with LiDAR enables RGB-LiDAR sensors that produce colored point clouds. Camera-based methods yield sparse features limited to high-gradient regions. Pure LiDAR captures dense geometry but lacks photometric texture. RGB-LiDAR combines both modalities, providing dense depth with per-point color for photorealistic 3D reconstruction and mapping.

State-of-the-art methods in 3D rendering such as 3D Gaussian Splatting~\cite{kerbl20233d} rely on point clouds extracted from images. Efforts to fuse LiDAR point clouds with images have emerged to obtain point clouds with richer texture. Multimodal data fusion has become increasingly important for high-fidelity 3D reconstruction, including structure-from-motion enhancement~\cite{zhen2020lidar}, neural radiance fields~\cite{tao2023lidarnerf}, object detection~\cite{liang2022deepfusion}, and Gaussian Splatting~\cite{xiong2024gauscene,lim2024lidar3dgs,patt2025densifybeforehand,chen2025lidargspp,strobel2025surffill,lindstrom2025idsplat,guo2025robust}. However, raw LiDAR scans present a significant challenge due to excessive data density. A single scan can contain millions of points, with substantial redundancy in both geometry and appearance. Standard downsampling methods---Random Sampling, Voxel Grid~\cite{rusu20113d}, and Normal Space Sampling~\cite{rusinkiewicz2001efficient} enforce spatial uniformity, discarding points indiscriminately in textured regions. Those regions are highly likely to contain useful features, while over-representing visually homogeneous surfaces like walls and floors. For applications requiring visual detail preservation, this approach is suboptimal for photometric tasks. It inadvertently discards high-frequency visual details required for realistic reproduction.

This motivates RGB color space as a stratification domain. \textit {We hypothesize that regions with high color variation contain visual detail worth preserving, while color-homogeneous regions represent photometric redundancy regardless of spatial extent.} By allocating sampling space proportional to color diversity rather than spatial area, we preserve texture essential for photometric reconstruction.

We introduce \textbf{PRISM (Point-cloud RGB indexed stratified sampling)}, a color-stratified sampling method for RGB-LiDAR point clouds. PRISM uses RGB color space as its strata and limits the number of points per color bin. This shifts sampling density from spatial uniformity to visual complexity, producing sparser point clouds that retain essential features for high-fidelity 3D reconstruction.

\section{Related Work}
\label{sec:related_work}

\noindent\textbf{Point Cloud Downsampling.} Since 3D point cloud data is often massive, it is preferred to sample a representative subset to reduce computational resources while retaining distinctive features of the scene. Among classical methods, random sampling is the simplest baseline~\cite{hu2020randlanet,guo2021deep,li2023comparative}. It is easy to implement and produces an unbiased subsample of the original point cloud, making it one of the most widely used approaches for large-scale point cloud processing. However, key essential features are reduced evenly with unimportant points. When unimportant regions dominate the point cloud, important features may be under-sampled.

To overcome this problem, stratified sampling strategies have been introduced. These methods partition the point cloud into several strata based on predefined criteria, then sample points within each stratum. Voxel-based sampling divides the space into a regular voxel grid (octree) as a stratum and samples representative points or centroids in each cell~\cite{li2023comparative,liu2020hierarchical,lyu2024dynamic}. While achieving uniform density, voxel-based methods are unaware to the photometric content of the scene. Normal Space Sampling (NSS) samples points using each point's normal vector, which is useful for retaining points closer to edges and curves~\cite{rusinkiewicz2001efficient,gelfand2003geometrically,kwok2019dnss}.

Recently, learning-based sampling strategies have been studied~\cite{dovrat2019learning,lang2020samplenet,ye2022apsnet}. However, these methods are not viable for whole scenes due to their computational cost and are focused mostly on object segmentation and classification within point clouds. Currently, no sampling techniques have incorporated colors as a stratum to sample point clouds. Unlike these methods, PRISM introduces color-aware selection, ensuring that points in texture-rich areas are preserved to better support the subsequent radiance field learning.

\noindent\textbf{3D Gaussian Splatting.} Recent advances in Novel View Synthesis (NVS) have been dominated by 3D Gaussian Splatting (3DGS)~\cite{kerbl20233d}, which represents scenes as a collection of anisotropic 3D Gaussians. Unlike coordinate-based neural networks, 3DGS relies heavily on a high-quality initial point cloud to anchor the optimization process. While SfM-derived points are the standard, they are often sparse and noisy. Many LiDAR-3DGS integration has been introduced, but most applied random and voxel based sampling to reduce LiDAR point cloud to be integrated to 3DGS. A notable work, LiDAR-3DGS demonstrates an early version of PRISM was demonstrated in LiDAR-3DGS~\cite{lim2024lidar3dgs}, which validated the effectiveness of color-based subsampling for 3D Gaussian Splatting. At 30k iterations, the method achieved a 7.064\% increase in PSNR and 0.564\% increase in SSIM compared to vanilla 3DGS. These results confirm that color-guided sampling preserves visually distinct features critical for photorealistic rendering. PRISM extends this approach by providing a more systematic framework for color stratification to enabling predictable compression ratios across diverse datasets.

\section{Preliminaries}
\label{sec:preliminaries}

\subsection{Camera-LiDAR Integration}

The fusion of LiDAR and RGB camera data requires precise calibration to establish correspondences between 3D geometric measurements and 2D image projections. A LiDAR sensor captures a point cloud $\mathcal{P} = \{p_i = (x_i, y_i, z_i)\}_{i=1}^N$ in its own coordinate frame, while an RGB camera observes the scene as pixel intensities $\mathcal{I}(u, v) = (r, g, b)$ in image coordinates. To generate a colored point cloud $\mathcal{P}_{\text{RGB}} = \{(x_i, y_i, z_i, r_i, g_i, b_i)\}_{i=1}^N$, each 3D point must be mapped to its corresponding pixel to retrieve RGB values.

This mapping relies on extrinsic calibration, which determines the rigid transformation between the LiDAR and camera coordinate systems. Let $\mathbf{R} \in SO(3)$ and $\mathbf{t} \in \mathbb{R}^3$ denote the rotation and translation from LiDAR to camera frame. A point $p_i = (x_i, y_i, z_i)$ in LiDAR coordinates is first transformed to camera coordinates:
\begin{equation}
p_i^{\text{cam}} = \mathbf{R} p_i + \mathbf{t}
\end{equation}

Given the camera intrinsic matrix $\mathbf{K}$ containing focal lengths $(f_x, f_y)$ and principal point $(c_x, c_y)$, the 3D point is projected onto the image plane:
\begin{equation}
\begin{bmatrix} u_i \\ v_i \\ 1 \end{bmatrix} \sim \mathbf{K} \begin{bmatrix} x_i^{\text{cam}} \\ y_i^{\text{cam}} \\ z_i^{\text{cam}} \end{bmatrix}
\end{equation}

The resulting pixel coordinates $(u_i, v_i)$ are used to retrieve color values $(r_i, g_i, b_i) = \mathcal{I}(u_i, v_i)$ from the RGB image, thereby associating photometric information with the geometric LiDAR measurement.

Both intrinsic and extrinsic calibration are essential for accurate correspondence. Misalignment between sensor coordinate systems results in color bleeding artifacts, where points receive incorrect RGB values from neighboring regions. In practice, intrinsic calibration determines the camera matrix $\mathbf{K}$ and distortion coefficients, while extrinsic calibration establishes the spatial relationship between sensors. Such calibration is often achieved through available ROS packages~\cite{ros2023camera,koide2023direct}, which enable robust alignment of visual and LiDAR data.

\subsection{Evaluation Metrics}

We evaluate downsampling quality using two standard geometric distance metrics. Let $\mathcal{P}$ denote the original point cloud and $\mathcal{P}_{\text{out}}$ the downsampled output.

\noindent\textbf{Chamfer Distance (CD)} computes the average bidirectional nearest-neighbor distance:
\begin{equation}
\text{CD}(\mathcal{P}, \mathcal{P}_{\text{out}}) = \frac{1}{|\mathcal{P}|} \sum_{p \in \mathcal{P}} \min_{q \in \mathcal{P}_{\text{out}}} \|p - q\| + \frac{1}{|\mathcal{P}_{\text{out}}|} \sum_{q \in \mathcal{P}_{\text{out}}} \min_{p \in \mathcal{P}} \|q - p\|
\end{equation}
This metric captures the average reconstruction error across both directions.

\noindent\textbf{Hausdorff Distance (HD)} measures the maximum nearest-neighbor distance:
\begin{equation}
\text{HD}(\mathcal{P}, \mathcal{P}_{\text{out}}) = \max \left\{ \sup_{p \in \mathcal{P}} \inf_{q \in \mathcal{P}_{\text{out}}} \|p - q\|, \sup_{q \in \mathcal{P}_{\text{out}}} \inf_{p \in \mathcal{P}} \|q - p\| \right\}
\end{equation}
This metric captures the worst-case error between the two point clouds.

Both metrics are widely used for evaluating point cloud downsampling methods~\cite{guo2021deep,lang2020samplenet}, as they quantify geometric fidelity without requiring point-to-point correspondence.

\section{Methodology}
\label{sec:methodology}

\subsection{Overview}

PRISM reduces RGB-LiDAR point cloud density while preserving visual detail through color-guided stratified sampling. Unlike spatial downsampling methods that treat all regions uniformly, PRISM uses color diversity as a proxy for visual importance. The algorithm bins points by RGB color and limits the number of points per bin. This preserves texture-rich regions with high chromatic variation while substantially reducing visually homogeneous regions.

\begin{figure}[t]
\centering
\includegraphics[width=\textwidth]{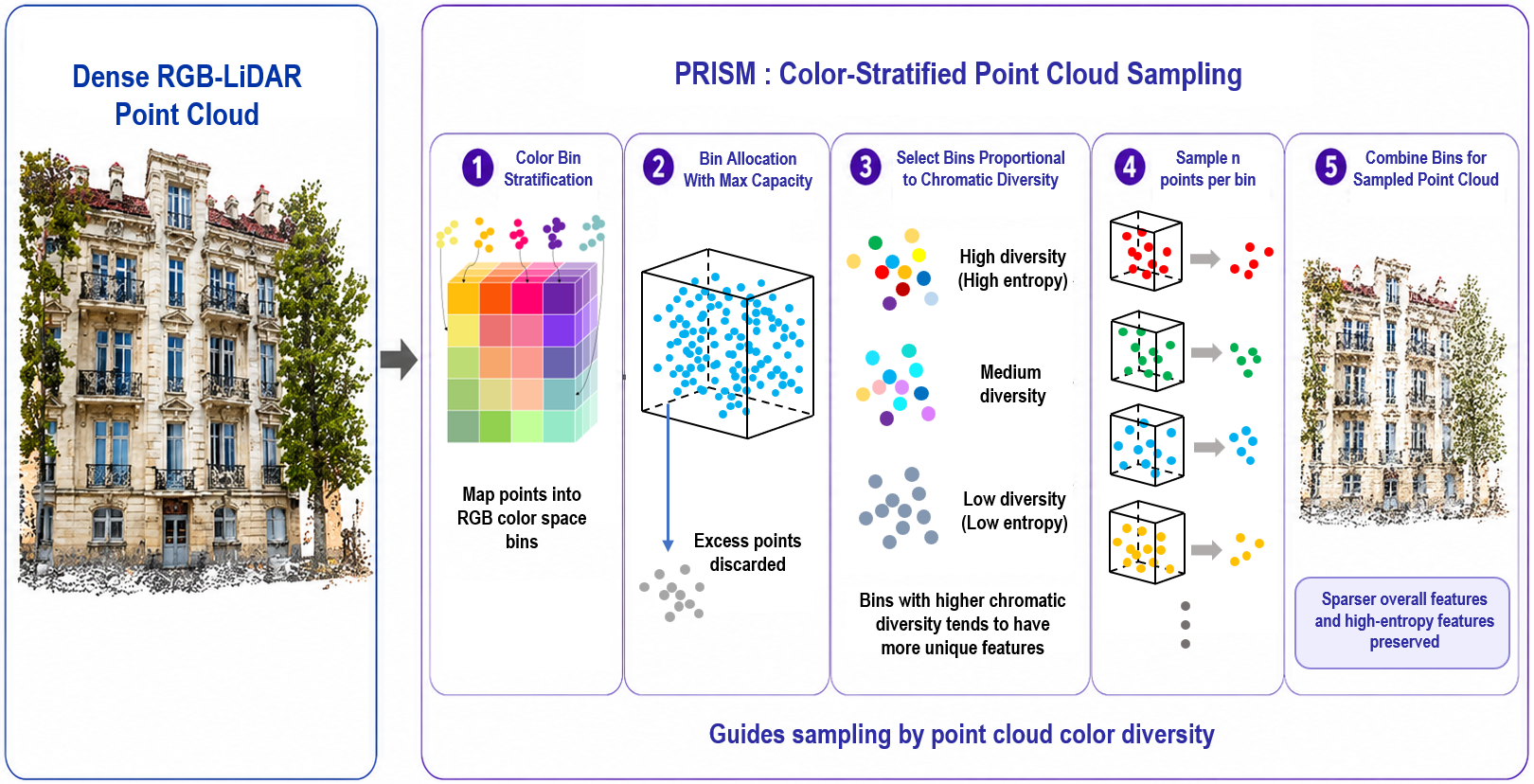}
\caption{Overview of PRISM. The input RGB-LiDAR point cloud is stratified in quantized RGB color space, capped by a globally selected per-bin capacity, and reconstructed as a downsampled point cloud that preserves chromatically diverse scene regions.}
\label{fig:prism_overview}
\end{figure}

Figure~\ref{fig:prism_overview} summarizes the PRISM pipeline. Given an RGB-LiDAR point cloud, PRISM indexes points by quantized RGB values, estimates a global bin capacity from the target compression ratio, and samples within each color stratum to produce a compact point cloud that emphasizes chromatically diverse regions.

\subsection{Color-based Stratification}

Let $\mathcal{P} = \{p_i\}_{i=1}^{N}$ be an input point cloud where each point $p_i = (\mathbf{x}_i, \mathbf{c}_i)$ consists of 3D coordinates $\mathbf{x}_i \in \mathbb{R}^3$ and RGB color $\mathbf{c}_i \in [0, 1]^3$. We treat RGB color space as the stratification domain. Colors are converted to integer bins:
\begin{equation}
    \mathbf{b}_i = \lfloor 255 \cdot \mathbf{c}_i \rfloor \in \{0, 1, \dots, 255\}^3,
\end{equation}
where each unique color bin $\mathbf{b} \in \mathcal{B}$ defines a stratum. Points sharing the same quantized RGB value are grouped together:
\begin{equation}
    \mathcal{G}_{\mathbf{b}} = \{p_i \mid \mathbf{b}_i = \mathbf{b}\}.
\end{equation}
This grouping partitions the point cloud into strata. Texture-rich regions yield many small groups with diverse colors, while homogeneous regions like walls and floors concentrate into few large groups.

\subsection{Stratified Sampling with Bin Limit}

The core sampling rule imposes a maximum capacity $k$ per color bin. For each stratum $\mathcal{G}_{\mathbf{b}}$, we retain at most $k$ points:
\begin{equation}
    \mathcal{S}_{\mathbf{b}} =
    \begin{cases}
        \text{Sample}(\mathcal{G}_{\mathbf{b}}, k) & \text{if } |\mathcal{G}_{\mathbf{b}}| > k, \\
        \mathcal{G}_{\mathbf{b}} & \text{otherwise},
    \end{cases}
\end{equation}
where $\text{Sample}(\mathcal{G}, k)$ randomly selects $k$ points from $\mathcal{G}$. The final downsampled point cloud is the union across all strata:
\begin{equation}
    \mathcal{P}_{\text{out}} = \bigcup_{\mathbf{b} \in \mathcal{B}} \mathcal{S}_{\mathbf{b}}.
\end{equation}
This allocates sampling density proportional to color diversity. A region with $M$ unique colors retains up to $M \cdot k$ points, while a single-color region retains at most $k$ points regardless of size.

\subsection{Global k Selection}

We specify a desired compression ratio $r_{\text{target}}$ rather than manually tuning $k$. PRISM determines the optimal $k$ to achieve this target. Let $n_{\mathbf{b}}$ denote the number of points in color bin $\mathbf{b}$. For a given cap $k$, the total output size is:
\begin{equation}
    S(k) = \sum_{\mathbf{b} \in \mathcal{B}} \min(n_{\mathbf{b}}, k).
\end{equation}
This function is monotonically increasing and piecewise linear in $k$. Our goal is to find $k^*$ such that $S(k^*) \approx r_{\text{target}} \cdot N$.

We exploit the piecewise linearity of $S(k)$. Let $n_1 \leq n_2 \leq \cdots \leq n_{|\mathcal{B}|}$ be the sorted bin sizes for some ordering of bins. Define cumulative sums $c_j = \sum_{i=1}^{j} n_i$ with $c_0 = 0$. For any $k$ in the interval $[n_j, n_{j+1}]$, the first $j$ bins contribute their full counts while the remaining $|\mathcal{B}| - j$ bins contribute $k$ each:
\begin{equation}
    S(k) = c_j + (|\mathcal{B}| - j) k.
\end{equation}
Setting $S(k) = r_{\text{target}} \cdot N$ yields a candidate solution:
\begin{equation}
    k_j^* = \frac{r_{\text{target}} \cdot N - c_j}{|\mathcal{B}| - j}, \quad j \in \{0, \ldots, |\mathcal{B}| - 1\}.
\end{equation}
We select the unique $j$ for which $n_j \leq k_j^* \leq n_{j+1}$, producing the real-valued solution $k^*$. The final integer cap is obtained by rounding and selecting the value closest to the target. This approach requires $O(|\mathcal{B}| \log |\mathcal{B}|)$ time dominated by sorting, which is asymptotically faster than binary search when $|\mathcal{B}| \ll N$. Once $k^*$ is determined, the sampling phase proceeds as described in the algorithm below.

\begin{center}
\small
\begin{tabular}{@{}p{0.95\columnwidth}@{}}
\toprule
Algorithm: PRISM \\
\midrule
Procedure \textsc{PRISM}($\mathcal{P}, r_{\text{target}}$) \\
\quad Input: Point cloud $\mathcal{P} = \{(\mathbf{x}_i, \mathbf{c}_i)\}_{i=1}^{N}$, target ratio $r_{\text{target}}$ \\
\quad Output: Downsampled point cloud $\mathcal{P}_{\text{out}}$ \\
1: Quantize colors: $\mathbf{b}_i \leftarrow \lfloor 255 \cdot \mathbf{c}_i \rfloor$ for all $p_i \in \mathcal{P}$ \\
2: Count points per bin: $n_{\mathbf{b}} \leftarrow |\{p_i \mid \mathbf{b}_i = \mathbf{b}\}|$ for all $\mathbf{b} \in \mathcal{B}$ \\
3: Solve for $k^*$ using the method in Section 3.4 \\
4: Group by color: $\mathcal{G}_{\mathbf{b}} \leftarrow \{p_i \mid \mathbf{b}_i = \mathbf{b}\}$ for all $\mathbf{b} \in \mathcal{B}$ \\
5: Initialize $\mathcal{P}_{\text{out}} \leftarrow \emptyset$ \\
6: for each stratum $\mathcal{G}_{\mathbf{b}}$ do \\
7: \quad if $|\mathcal{G}_{\mathbf{b}}| > k^*$ then \\
8: \quad \quad Randomly sample $k^*$ points: $\mathcal{S}_{\mathbf{b}} \leftarrow \text{Sample}(\mathcal{G}_{\mathbf{b}}, k^*)$ \\
9: \quad else \\
10: \quad \quad Keep all points: $\mathcal{S}_{\mathbf{b}} \leftarrow \mathcal{G}_{\mathbf{b}}$ \\
11: \quad $\mathcal{P}_{\text{out}} \leftarrow \mathcal{P}_{\text{out}} \cup \mathcal{S}_{\mathbf{b}}$ \\
12: end for \\
13: return $\mathcal{P}_{\text{out}}$ \\
End Procedure \\
\bottomrule
\end{tabular}
\end{center}

\section{Results and Analysis}

\subsection{Quantitative Results}

We evaluate PRISM on three datasets that represent distinct scene characteristics: Toronto-3D~\cite{Toronto3D} captures urban outdoor environments with mobile LiDAR, providing diverse structures and natural color variation. ETH3D~\cite{ETH3D} offers high-resolution architectural scenes with structured color patterns. Paris-CARLA~\cite{ParisCARLA} provides synthetic urban scenes with rich textures and high chromatic diversity. All experiments were conducted on a system with AMD Ryzen 9 9800X3D CPU and 64GB RAM.

Table \ref{tab:main_results} presents the quantitative comparison of PRISM against baseline sampling strategies on these datasets. We compare Random Sampling, Voxel Grid downsampling, Normal Space Sampling, and our method at target compression ratios near 1\%. For PRISM, the bin capacity $k$ is automatically selected per dataset to achieve the target ratio (see Section 4.4), yielding $k=1$ for all three datasets.

\begin{table*}[t]
\centering
\caption{Quantitative comparison at target compression ratio $\approx$1\%. \textbf{CD}: Chamfer Distance ($\downarrow$), \textbf{HD}: Hausdorff Distance ($\downarrow$), \textbf{Entropy}: Color Entropy Gain ($\uparrow$).}
\label{tab:main_results}
\begin{tabular}{l|l|c|cc|c|r}
\toprule
\textbf{Method} & \textbf{Dataset} & \textbf{Ratio (\%)} & \textbf{CD ($\downarrow$)} & \textbf{HD ($\downarrow$)} & \textbf{Entropy ($\uparrow$)} & \textbf{Time (s)} \\
\midrule
\multirow{3}{*}{Random} & Toronto-3D & 0.97 & \textbf{0.49} & 58.4 & -0.23 & \textbf{67.1} \\
 & ETH3D & 1.00 & \textbf{0.18} & \textbf{363.9} & -0.08 & \textbf{45.0} \\
 & Paris-CARLA & 0.97 & \textbf{0.71} & \textbf{49.3} & -0.12 & \textbf{302.8} \\
\midrule
\multirow{3}{*}{Voxel Grid} & Toronto-3D & 5.90 & 0.75 & \textbf{53.6} & 1.48 & 78.3 \\
 & ETH3D & 0.66 & 0.46 & 447.0 & 0.39 & 49.7 \\
 & Paris-CARLA & 1.30 & 0.99 & 43.1 & 1.43 & 328.8 \\
\midrule
\multirow{3}{*}{Normal Space} & Toronto-3D & 1.51 & 0.62 & 59.3 & 1.33 & 92.3 \\
 & ETH3D & 1.39 & 0.50 & 405.9 & 1.32 & 59.4 \\
 & Paris-CARLA & 0.30 & 2.07 & 47.2 & 1.97 & 367.3 \\
\midrule
\multirow{3}{*}{\textbf{PRISM}} & Toronto-3D & \textbf{1.37} & 0.83 & 58.3 & \textbf{3.08} & 73.4 \\
 & ETH3D & \textbf{0.90} & 0.35 & 366.4 & \textbf{2.20} & 50.5 \\
 & Paris-CARLA & \textbf{1.19} & 1.53 & 45.6 & \textbf{3.82} & 422.9 \\
\bottomrule
\end{tabular}
\end{table*}

\noindent\textbf{Compression Ratio Control.}
A critical challenge in point cloud downsampling is achieving predictable compression ratios. As shown in Table \ref{tab:main_results}, PRISM consistently achieves ratios close to the 1\% target (1.37\%, 0.90\%, 1.19\%) across all datasets through the global k selection. The bin capacity $k$ adapts to each scene's color complexity. 

In contrast, Voxel Grid downsampling exhibits significant variability. on Toronto-3D, it retains 5.90\% of points, while on ETH3D it retains only 0.66\%. This unpredictability stems from Voxel Grid's dependence on spatial density distribution. Random Sampling achieves stable compression ratios (0.97--1.00\%), but at the cost of indiscriminate point removal. PRISM combines the stability of Random Sampling with color-stratified selection, adapting to scene complexity while maintaining predictable output size.

\noindent\textbf{Geometric Accuracy Trade-off.}
PRISM exhibits higher Chamfer Distance than Random Sampling across all datasets. This reflects the method's design priority. It preserves color-diverse regions at the expense of uniform spatial coverage. On the ETH3D dataset, PRISM achieves CD of 0.35m, which is competitive with geometry-based methods (Voxel Grid: 0.46m, Normal Space: 0.50m) despite prioritizing chromatic rather than geometric features. This suggests that color stratification can serve as a reasonable proxy for spatial importance in certain environments, particularly architectural scenes with structured color variation.

On the Paris-CARLA dataset, PRISM shows higher geometric error (CD: 1.53m) compared to baselines. This dataset contains rich synthetic textures with high chromatic diversity, causing PRISM to substantially oversample visually distinct regions while under-sampling large homogeneous surfaces. While this increases geometric error, it preserves semantic structures (e.g., building facades, street furniture) that might be useful for downstream tasks like 3D Gaussian Splatting, where visual fidelity outweighs millimeter-level geometric precision.

\noindent\textbf{Computational Efficiency.}
PRISM demonstrates competitive runtime performance on Toronto-3D (73.4s) and ETH3D (50.5s), comparable to Random Sampling and faster than Normal Space Sampling. The algorithm's computational cost scales linearly with the number of input points and the number of unique color bins. On Paris-CARLA, runtime increases to 422.9s due to the dataset's high chromatic complexity, which creates more color bins and requires more sampling operations.

\noindent\textbf{Color Entropy Preservation.}
Color entropy measures the diversity of the color distribution in a point cloud. Negative entropy gain indicates that downsampling reduces chromatic diversity by disproportionately removing rare colors. PRISM consistently achieves the highest color entropy gain across all datasets: Toronto-3D (3.08), ETH3D (2.20), and Paris-CARLA (3.82). Random Sampling exhibits negative entropy gain on all datasets (-0.23, -0.08, -0.12), confirming loss of color variation through indiscriminate removal. Geometry-based methods preserve intermediate entropy levels (Voxel Grid: 0.39-1.48, Normal Space: 1.32-1.97) but remain substantially lower than PRISM.

\subsection{Qualitative Analysis}

\begin{figure}[b]
\centering
\includegraphics[width=0.48\columnwidth]{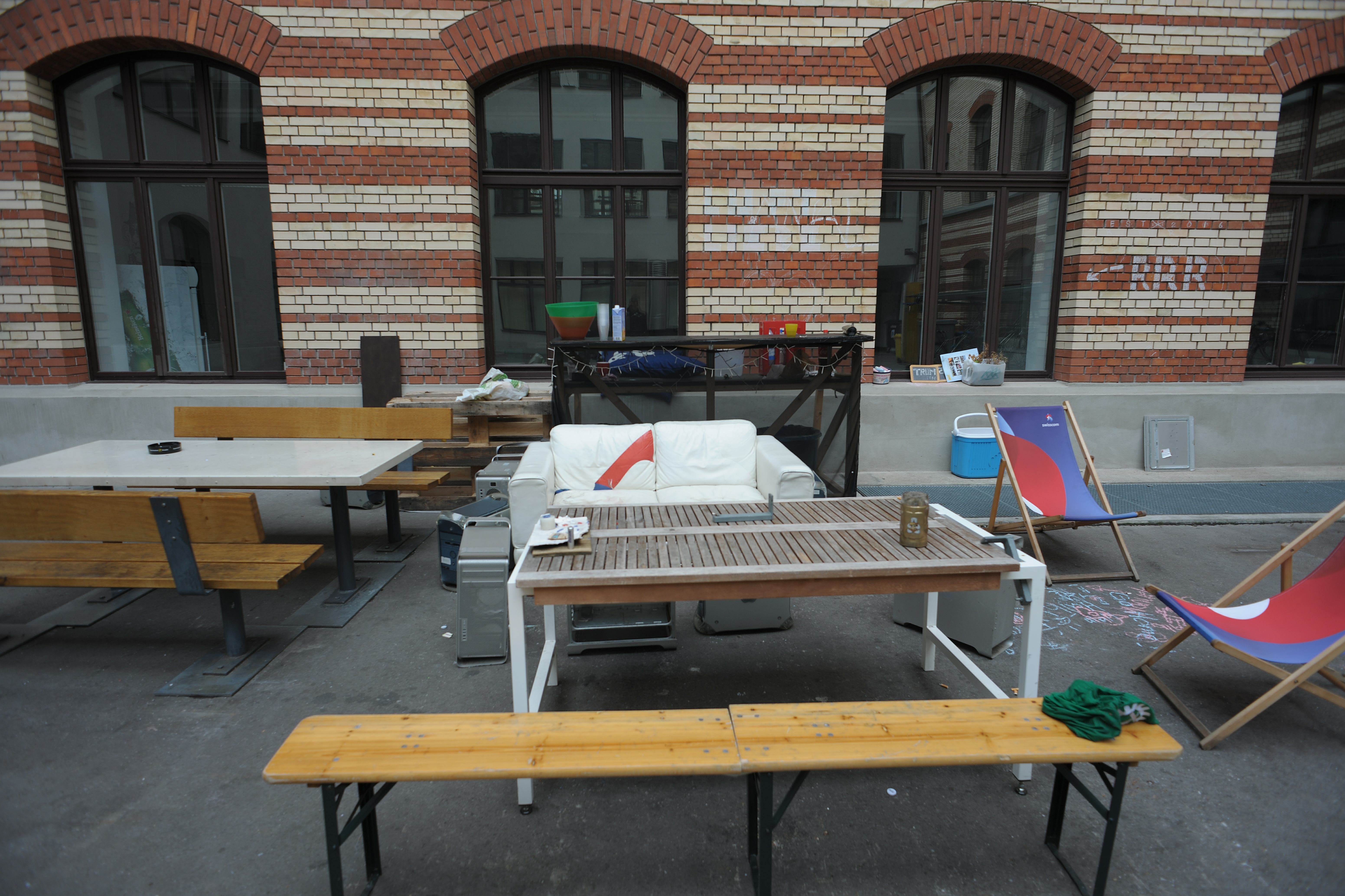}
\hfill
\includegraphics[width=0.48\columnwidth]{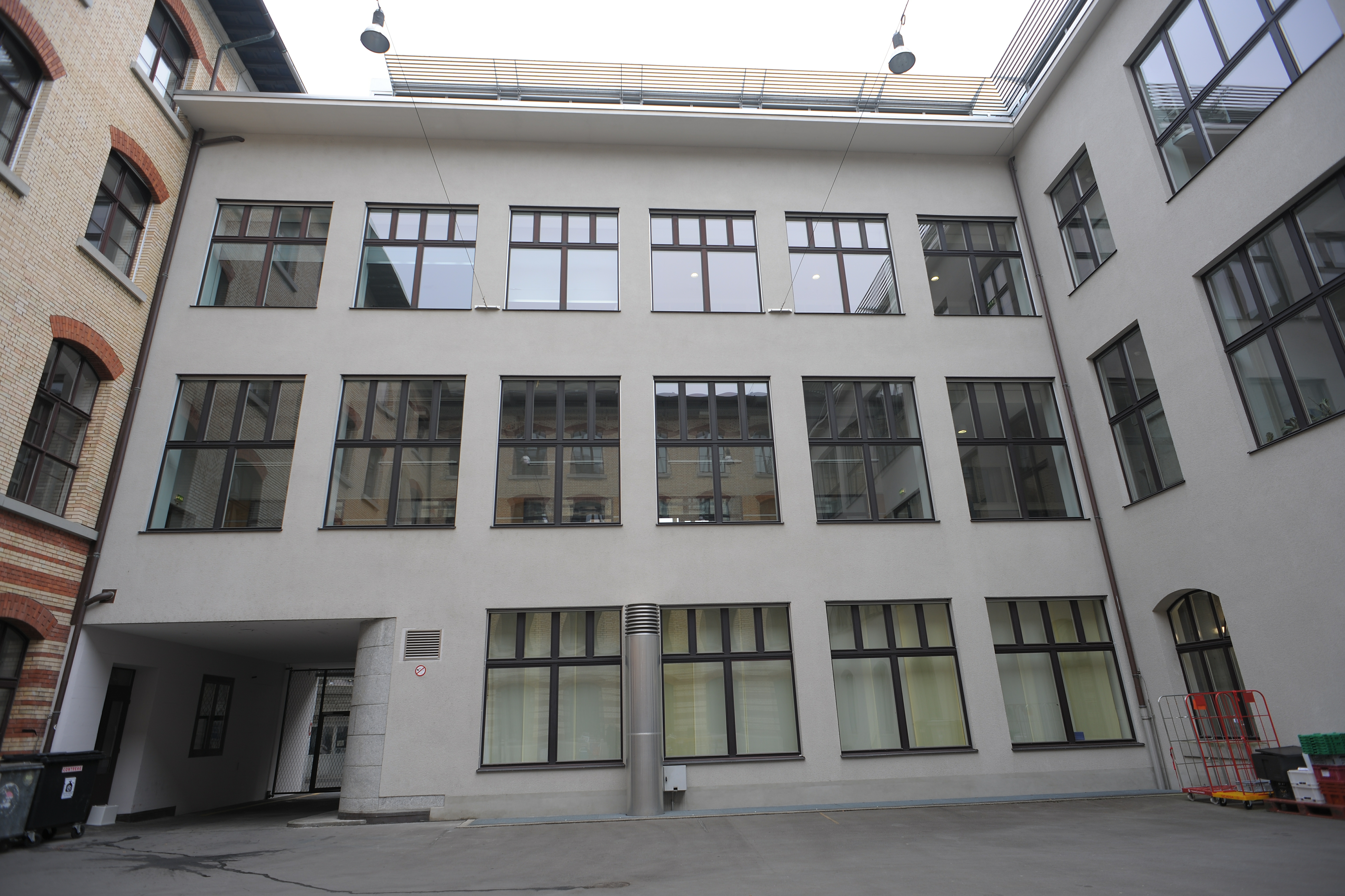}
\caption{ETH3D courtyard scene. The scene contains high color variations in small objects present in the scene alongside building walls with same color.}
\label{fig:scene}
\end{figure}

We examine PRISM's behavior on the ETH3D courtyard scene, which exhibits high color diversity in architectural details while containing large photometrically homogeneous regions such as walls and floors. The compact spatial extent of this scene makes it suitable for detailed visualization. Figure \ref{fig:scene} shows representative views of the scene. This combination of chromatic complexity and spatial redundancy makes it an ideal test case for color-guided sampling.

\begin{figure*}[!t]
\centering
\includegraphics[width=0.8\columnwidth]{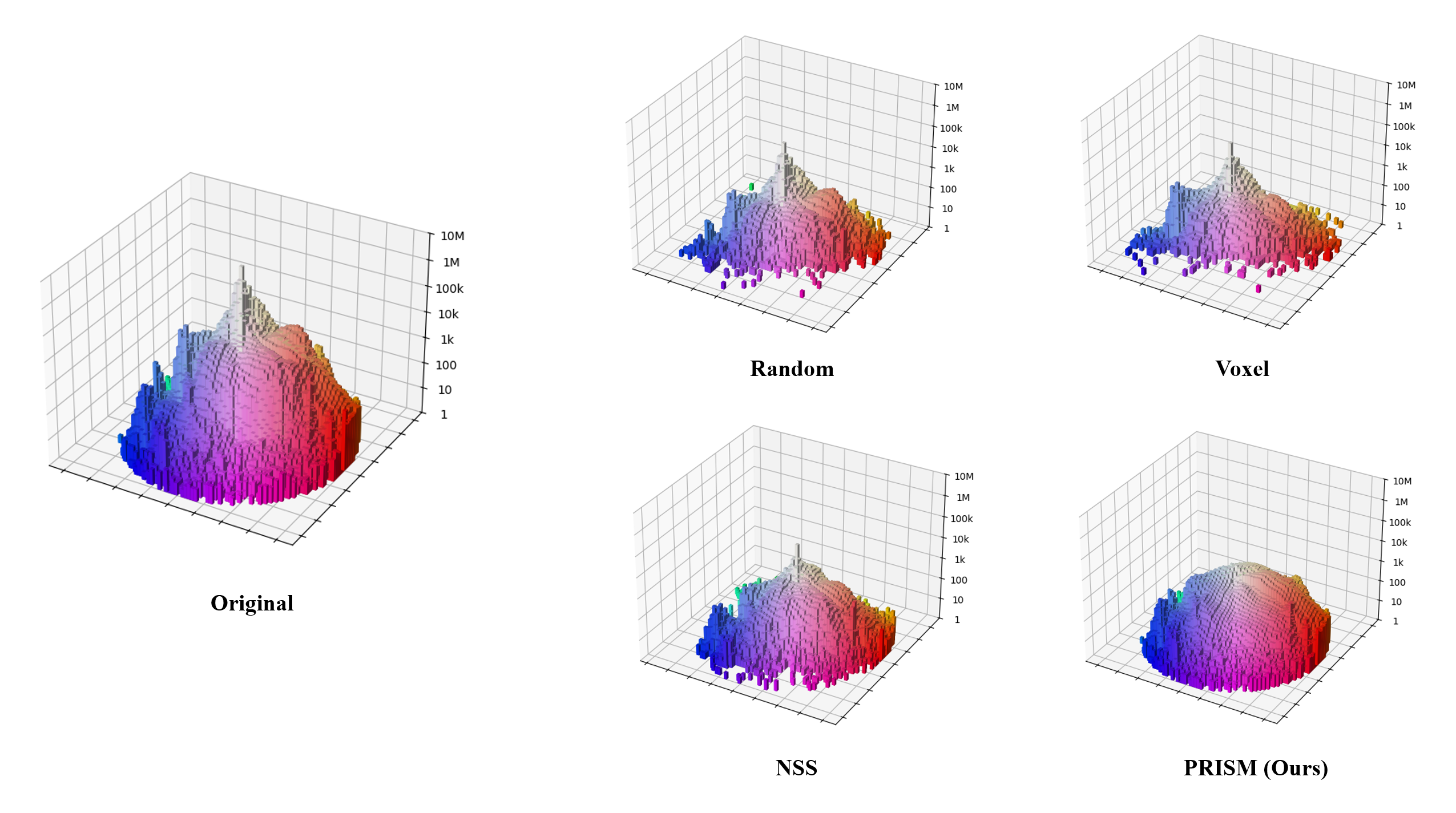}
\caption{Color distribution comparison of PRISM and baseline methods on the courtyard scene. PRISM preserves chromatic diversity closely aligned with the input distribution. All of the baselines were sampled close to 1\%}
\label{fig:histogram_comparison}
\end{figure*}

\begin{figure*}[!b]
\centering
\includegraphics[width=0.80\columnwidth]{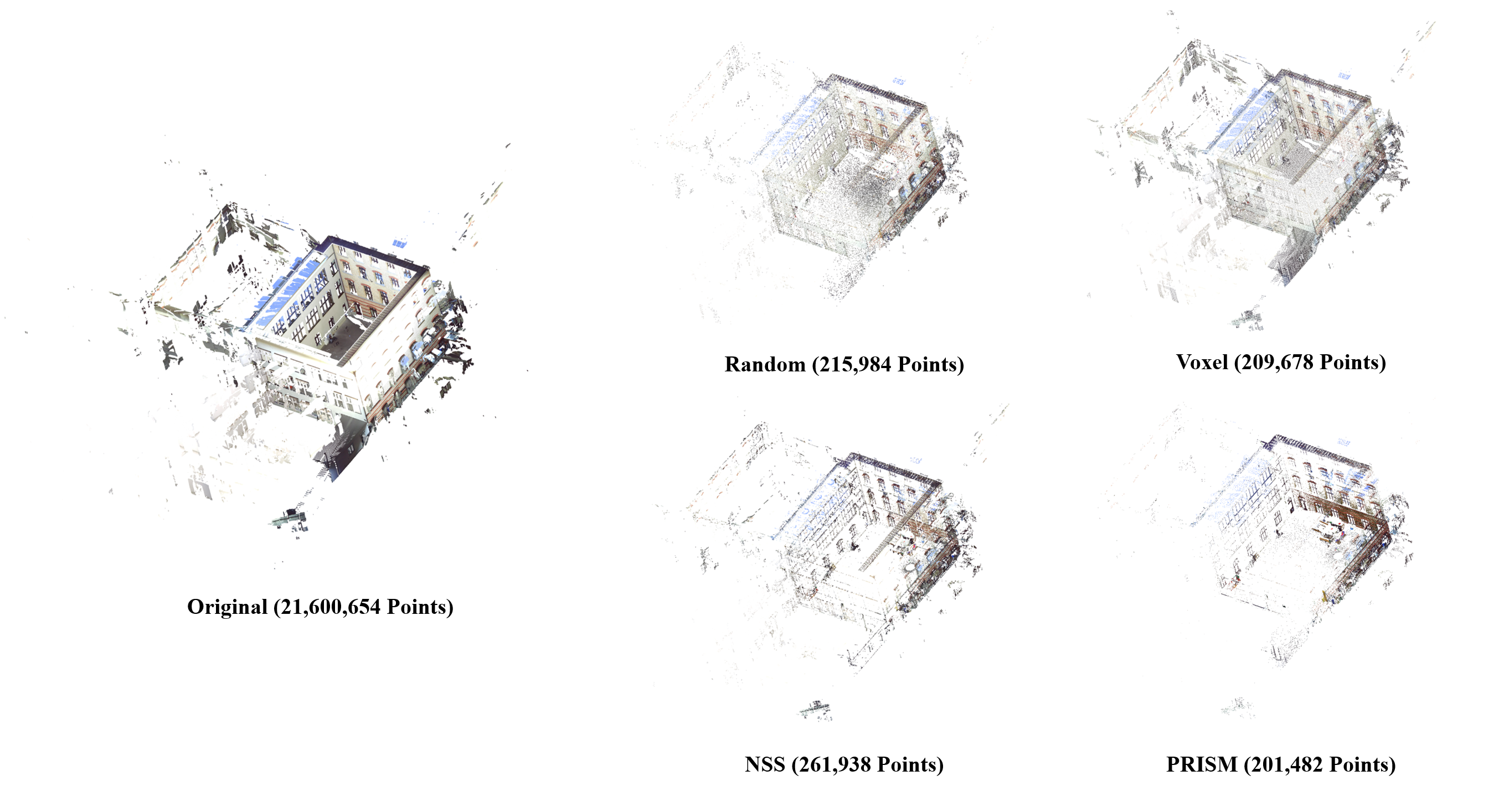}
\caption{Point cloud comparison of PRISM and baseline methods. PRISM retains higher density in texture-rich regions with chromatic variance.}
\label{fig:pointcloud_comparison}
\end{figure*}

\noindent\textbf{Color Distribution Analysis.}
We visualize color distributions using a polar chroma histogram in HSV space. RGB values are converted to cylindrical coordinates $(r, \theta, z)$, where saturation maps to $r$, hue to $\theta$, and bin count to $z$. The $(r, \theta)$ plane is discretized into a fixed grid. Bin heights are shown on a logarithmic scale with a shared global range for consistent comparison.

PRISM's per-bin cap $k$ produces a flattened vertical distribution (Fig. \ref{fig:histogram_comparison}). In contrast, baseline methods preserve strong height disparities from overrepresented colors while removing rarer colors which may correspond to unique features of the scene. This demonstrates that PRISM limits color concentration while maintaining chromatic coverage, including rare colors.

Figure \ref{fig:histogram_comparison} compares color distributions on the ETH3D courtyard scene. The scene is dominated by grey building walls (Fig. \ref{fig:scene}), creating strong grey concentration in the input. Baseline methods exhibit bias toward dominant grey, oversampling redundant walls while discarding unique features. In contrast, PRISM substantially downsamples grey points while preserving rare colors that correspond to unique scene features, demonstrating effective photometric redundancy suppression.

\noindent\textbf{Feature Preservation.}
Figure \ref{fig:pointcloud_comparison} visualizes the effect of color-guided sampling. In texture-rich regions such as window frames, signage, and architectural ornaments, PRISM maintains higher point density compared to geometry-based methods. These features, while geometrically small, carry high chromatic variance and can be useful for photorealistic rendering where point cloud diversity directly influences splat initialization. Voxel Grid downsampling allocates points based on spatial occupancy, causing large homogeneous regions to receive dense sampling despite photometric redundancy. Normal Space Sampling prioritizes geometric variation but remains agnostic to color diversity. This confirms that color stratification effectively identifies regions of photometric importance.

\begin{figure*}[!b]
\centering
\includegraphics[width=0.80\columnwidth]{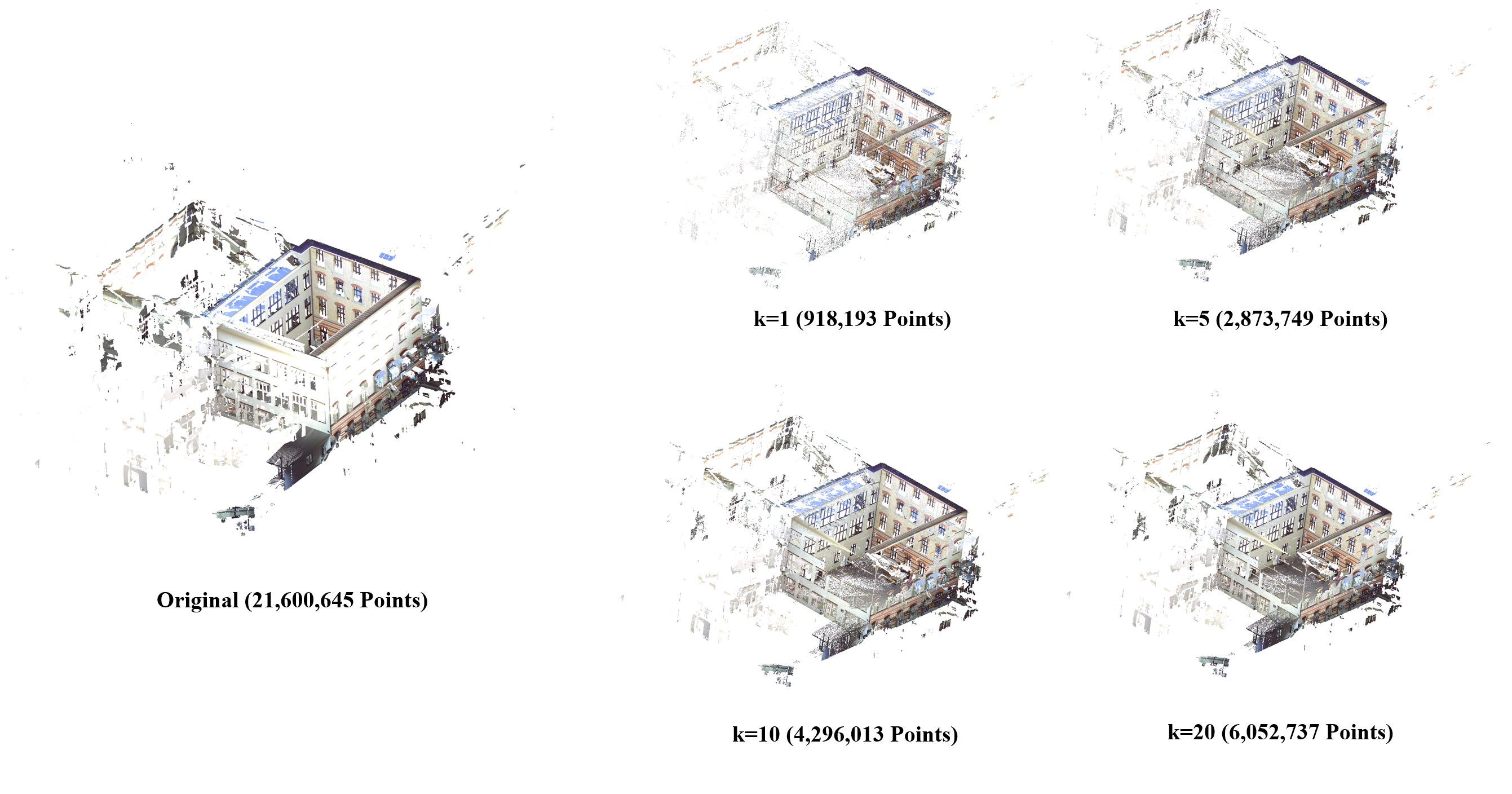}
\caption{Effect of bin capacity $k$ on point cloud density. As $k$ increases, compression ratio grows while maintaining color-guided sampling characteristics.}
\label{fig:k_sensitivity}
\end{figure*}

\noindent\textbf{Bin Capacity Effect.}
Figure \ref{fig:k_sensitivity} illustrates the effect of bin capacity $k$ on point cloud density. As $k$ increases from 1 to 20, the output grows from 918k to 6.05M points (4.25\% to 28.0\% of the original 21.6M points). At $k=1$, extreme sparsity preserves only the most chromatically diverse regions, retaining one representative point per unique color. At $k=10$, the method retains 4.3M points (19.9\%), balancing compression and spatial coverage. Higher values like $k=20$ increase density further while maintaining color-guided selection. The visual comparison shows that PRISM consistently preserves texture-rich architectural details across all $k$ values, with density scaling proportionally to the parameter. This demonstrates predictable compression control through a single intuitive parameter.


\subsection{Ablation Study}

PRISM's design involves two key parameters: chromaticity normalization and quantization level. Chromaticity normalization removes luminance information, merging points with identical hue but different brightness caused by varying illumination into the same bin. Quantization groups adjacent color values together, reducing the total number of color bins by treating similar colors as identical. Table \ref{tab:ablation} examines how these parameters affect compression ratio and geometric accuracy.

\begin{table}[!b]
\centering
\caption{Ablation study on ($k=10$).}
\label{tab:ablation}
\begin{tabular}{c|l|c|cc|c|r}
\toprule
\textbf{Quant.} & \textbf{Chromaticity} & \textbf{Ratio (\%)} & \textbf{CD ($\downarrow$)} & \textbf{HD ($\downarrow$)} & \textbf{Entropy ($\uparrow$)} & \textbf{Time (s)} \\
\midrule
\multirow{2}{*}{1-bit} & w/ Chromaticity & 0.77 & 0.44 & 456.0 & 2.65 & 45.5 \\
 & w/o Chromaticity & 17.50 & \textbf{0.27} & \textbf{410.6} & 2.80 & 60.6 \\
\midrule
\multirow{2}{*}{2-bit} & w/ Chromaticity & 0.24 & 0.56 & 369.6 & 2.14 & 52.9 \\
 & w/o Chromaticity & 4.83 & 0.30 & 365.8 & 3.65 & 57.5 \\
\midrule
\multirow{2}{*}{4-bit} & w/ Chromaticity & 0.10 & 0.75 & 371.3 & 1.58 & 51.7 \\
 & w/o Chromaticity & 1.29 & 0.35 & 324.1 & 4.13 & 51.4 \\
\bottomrule
\end{tabular}
\end{table}

Table \ref{tab:ablation} presents ablation results, comparing the effect of chromaticity normalization and quantization levels. The parameter $k$ is fixed at 10 for this analysis.

\medskip
\medskip

\noindent\textbf{Chromaticity Normalization.}
The choice between w/ Chromaticity and w/o Chromaticity represents a fundamental trade-off between compression aggressiveness and geometric fidelity. Table \ref{tab:ablation} shows that w/ Chromaticity achieves high compression (0.77\% at 1-bit) with higher Chamfer Distance (0.44m). This occurs because luminance variations carry geometric information---lighting changes across corners, edges, and curved surfaces provide shading cues. Discarding luminance merges these geometrically distinct points, reducing spatial coverage.

In contrast, w/o Chromaticity preserves the original RGB values, treating colors with different brightness levels as distinct bins. This results in far more bins and thus higher retention ratios (17.50\% at 1-bit). While this preserves geometric accuracy (CD: 0.27m), it reduces compression effectiveness. Chromaticity normalization is therefore recommended when the goal is very high compression with hue-based diversity preservation, while w/o Chromaticity is better suited for moderate compression with photometric fidelity.

\noindent\textbf{Quantization Level.}
Increasing the quantization level (1-bit $\rightarrow$ 2-bit $\rightarrow$ 4-bit) reduces the number of distinct color bins by grouping adjacent color values. Higher quantization merges more colors into fewer bins, concentrating points into a smaller number of strata. Since each bin is capped at $k=10$ points, this concentration increases the number of points discarded per bin. For w/ Chromaticity, the ratio decreases from 0.77\% (1-bit) to 0.10\% (4-bit), demonstrating effective compression control through quantization.

However, higher quantization comes at the cost of spatial uniformity. At 4-bit w/ Chromaticity, fewer bins force spatially distant points with similar chromaticity to compete for the same $k=10$ slots, reducing spatial coverage. 

This results in higher Chamfer Distance (0.75m vs 0.44m at 1-bit), as retained points are less representative of the full geometry. This suggests that 1-bit or 2-bit quantization offers the best balance between compression and geometric coherence for most applications.

\noindent\textbf{Color Entropy Dynamics.}
Table \ref{tab:ablation} reveals distinct entropy trends. For w/ Chromaticity, entropy gain declines as quantization increases (2.65 $\rightarrow$ 1.58). Since luminance is already removed, coarser quantization merges spectrally distinct hues, reducing the diversity of the retained palette. In contrast, w/o Chromaticity shows increased entropy gain with coarser quantization (2.80 $\rightarrow$ 4.13). By grouping spatially abundant but slightly varying background colors (e.g., noisy walls) into fewer bins, coarser quantization allows the capacity limit $k$ to suppress dominant regions more effectively, thereby increasing the relative frequency of rare, information-rich colors.

\section{Conclusion}

We presented PRISM, a color-guided stratified sampling method for RGB-LiDAR point clouds. By treating RGB color space as the stratification domain and imposing a per-bin capacity constraint, PRISM shifts sampling space from spatial uniformity to chromatic diversity. This approach preserves rare colors that correspond to unique scene features while substantially reducing repetitive and redundant features in homogeneous regions.

Our hypothesis---that regions with high color variation contain visual detail worth preserving, while color-homogeneous regions represent photometric redundancy---is validated by experimental results across three diverse datasets. PRISM achieves consistent compression ratios near 1\% while maintaining the full chromatic spectrum, including rare colors that correspond to unique scene features. Qualitative analysis on the ETH3D courtyard scene demonstrates that color stratification successfully identifies texture-rich regions such as window frames and architectural ornaments. These regions receive higher point density where photometric detail is concentrated. Conversely, large homogeneous surfaces like walls and floors are substantially downsampled. This confirms that chromatic uniformity can serve as a proxy for photometric redundancy.

The ablation study reveals trade-offs in parameter selection: chromaticity normalization enables extreme compression by merging illumination variations, while quantization controls the granularity of color stratification. PRISM exhibits higher geometric error compared to spatial methods, reflecting the prioritization of visual fidelity over uniform spatial coverage. For applications requiring millimeter-level geometric accuracy, spatial methods remain preferable.

Future work will investigate PRISM's impact on downstream applications, particularly SLAM systems and novel view synthesis. Additionally, combining the advantages of color-based diversity preservation with spatial regularization techniques could yield improved geometric accuracy while maintaining visual fidelity. Understanding how color-guided sampling affects these reconstruction pipelines remains an important research direction.

%
%
\paragraph{Acknowledgments.}
This work was supported by the National Research Foundation of Korea (NRF) grant funded by the Korea government (MSIT) (Grant No. RS-2022-NR067080 and RS-2025-05515607).

%
%
\bibliographystyle{splncs04}
\bibliography{main}
\end{document}